\documentclass[conference]{IEEEtran}
\usepackage{blindtext, graphicx}
\usepackage{mathtools}
\usepackage[utf8]{inputenc}
\usepackage[english]{babel}
\usepackage{comment}
\usepackage{csquotes}
\usepackage{hyperref}

\usepackage[backend=bibtex,style=ieee,sorting=ynt]{biblatex}
\ifCLASSINFOpdf
\else
\fi

\begin{document}
\title{Radio Transformer Networks: Attention Models for Learning to Synchronize in Wireless Systems}

\author{\IEEEauthorblockN{Timothy J. O'Shea}
\IEEEauthorblockA{Virginia Tech\\
Arlington, VA\\
oshea@vt.edu}
\and
\IEEEauthorblockN{Latha Pemula}
\IEEEauthorblockA{Virginia Tech\\
Blacksburg, VA\\
lpemula@vt.edu}
\and
\IEEEauthorblockN{Dhruv Batra}
\IEEEauthorblockA{Virginia Tech\\
Blacksburg, VA\\
dbatra@vt.edu}
\and
\IEEEauthorblockN{T. Charles Clancy}
\IEEEauthorblockA{Virginia Tech\\
Arlington, VA\\
tcc@vt.edu}
}

\maketitle

\begin{abstract}
We introduce learned attention models into the radio machine learning domain for the task of modulation recognition by leveraging spatial transformer networks and introducing new radio domain appropriate transformations.  This attention model allows the network to learn a localization network capable of synchronizing and normalizing a radio signal blindly with zero knowledge of the signal's structure based on optimization of the network for classification accuracy, sparse representation, and regularization.  Using this architecture we are able to outperform our prior results in accuracy vs signal to noise ratio against an identical system without attention, however we believe such an attention model has implication far beyond the task of modulation recognition.
\end{abstract}


\IEEEpeerreviewmaketitle

\section{Introduction}

Cognitive radio and signal processing in general has long relied on a relatively well defined set of expert systems and expert knowledge to operate.  Unfortunately in the realization of cognitive radio, this has greatly limited the ability of systems to generalize and perform real learning and adaptation to new and unknown signals and tasks.  By approaching signal recognition, synchronization, and reasoning from a ground-up feature learning angle, we seek to be able to build cognitive radio systems which truly generalize and adapt without running into barriers of expert knowledge such as many of the current day solutions which address more narrowly scoped problems.

In our prior work, we looked at the application of deep convolutional neural networks to the task of modulation recognition \cite{convmodrec} through blind feature learning on time domain radio signals.  We were able to achieve excellent classification performance at both low and high SNR by learning time domain features directly from a dataset with harsh channel impairments (oscillator drift, clock drift, fading, noise).  However we had no notion of attention in this work and instead forced the discriminative network to learn features invariant to each of these channel effects.  In communications receivers (and many iterative expert modulation classification algorithms), we typically perform synchronization on the signal before performing additional signal processing steps.  This synchronization can be thought of as a form of attention which estimates a time, frequency, phase, and sample timing offset in order to create a normalized version of the signal.

Attention models have recently been gaining widespread adoption in the computer vision community for a number of important reasons.  They introduce a learned model for attention capable of removing numerous variances and parametric search spaces in the input data and focuses on the task of extracting a canonical form attention patch with these variations removed to make downstream tasks easier and of lower complexity.  These were first introduced as recurrent networks \cite{mnih2014recurrent} which were quite expensive, but have made significant progress since then.

\begin{figure}[ht!]
  \centering
      \includegraphics[width=0.5\textwidth]{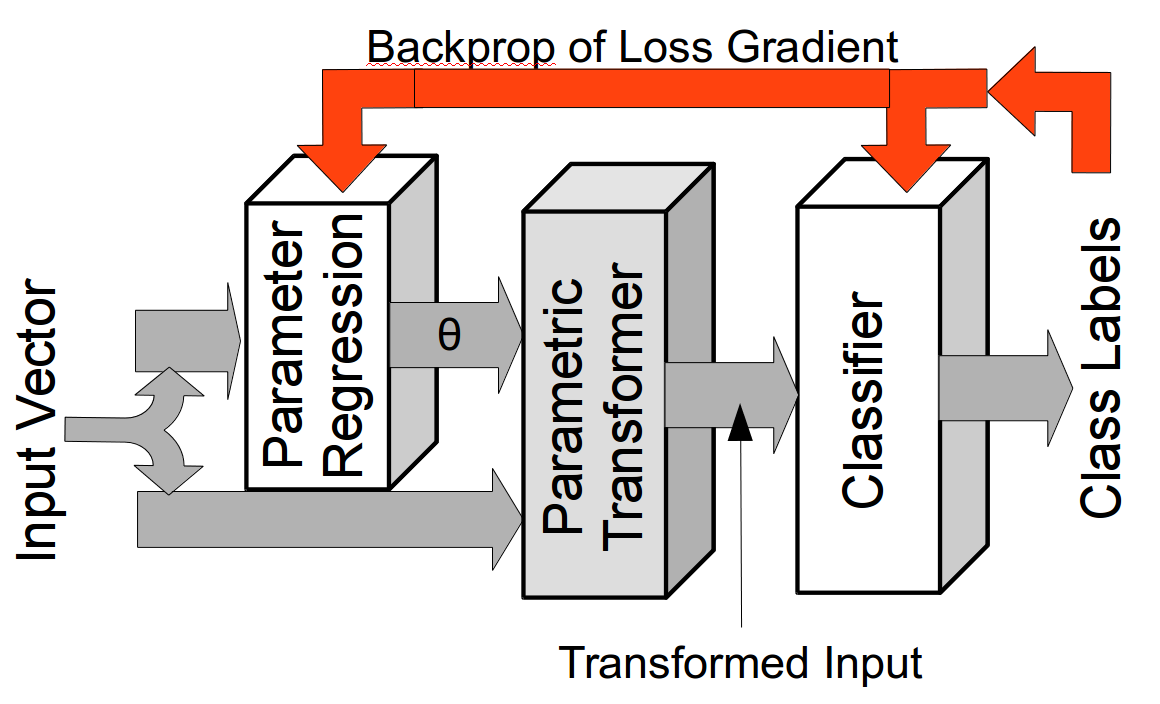}
        \caption{Generalized Transformer Network Architecture}
    \label{fig:stn}
\end{figure}

Spatial transformer networks (STNs) \cite{jaderberg2015spatial} were recently introduced to provide an end to end feed-forward model of attention which can be trained directly from loss on each training example and compactly evaluated on new samples.  They consist of a trained Localization network which performs parameter regression, a fixed parametric transform operation, and a trained discriminative classifier to select a class estimate.  In the image domain, where these have so far been applied, a 2D Affine transform is used to extract an attention patch which is shifted, scaled, and rotated in 2-space from the original image according to a 2x3 parameter vector $\theta$, generalized in figure \ref{fig:stn}.

In this work, we propose a Radio Transformer Network (RTN), which leverages the generalization of the STN architecture, but introduces radio-domain specific parametric transforms.  This attention model can be used to learn directly how to synchronize in wireless systems, and enables our modulation recognition system to outperform the attention-less version of itself by assisting in the normalization of the received signal prior to classification.  By constructing this normalized received signal with an attention model we greatly simplify the task of the discriminative network by relaxing the requirements on various variations of the received signal it must recognize, reducing the complexity and increasing the performance necessary in a discriminative network.  

This is an important result in modulation recognition but also more widely in radio communications and signal processing, as it demonstrates that we can \textbf{learn to synchronize} rather than relying on expert systems and estimators derived through a costly analytic process.  We believe attention models will play an important and wide-spread role in forthcoming machine learning based signal processing systems.

\section{Learning to Classify Signals}

\begin{figure}[ht!]
  \centering
      \includegraphics[width=0.5\textwidth]{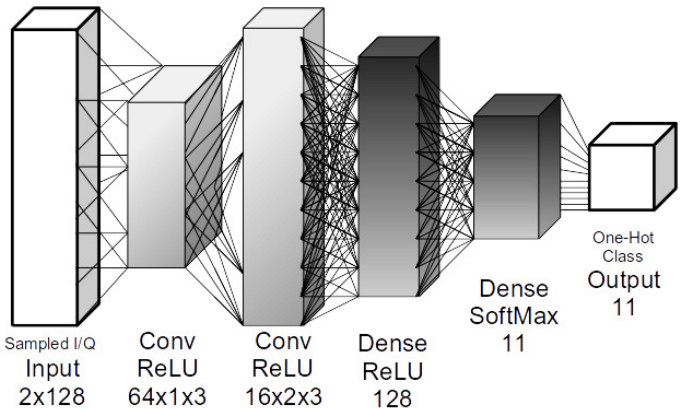}
        \caption{Original ConvNet Architecture without Attention}
    \label{fig:arch1}
\end{figure}

In prior work \cite{convmodrec} we compare supervised learning using a deep convolutional network with no expert features, to a handful of widely used machine learning techniques on expert signal amplitude, phase, and envelope moments.  We used the architecture shown in figure \ref{fig:arch1} using a convolutional frontend, a dense backend, and a softmax with categorical cross-entropy training using Adam/SGD against a synthetic dataset.

\begin{figure}[ht!]
  \centering
      \includegraphics[width=0.5\textwidth]{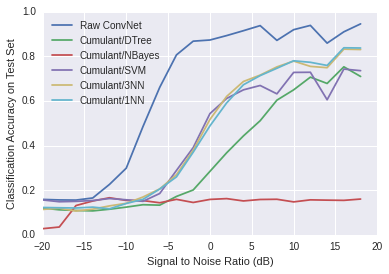}
        \caption{Original ConvNet Performance vs Expert Statistics}
    \label{fig:perf1}
\end{figure}

In figure \ref{fig:perf1} we summarize the results of that experiment without an attention model, demonstrating a significant improvement over the moment based features and conventional classifiers.

This is an exciting result as it demonstrates that feature learning on raw RF data does work, and in this case it is working better than conventional widely used expert features!

\section{Using Attention to Synchronize}

To effectively synchronize to a wireless signal, we must develop a transform which, with the right parameters is able to correct for channel induced variation.  Within the scope of this paper, we consider channel variation due to time offset, time dilation, frequency offset, and phase offset.  These are effects which exist in any real system containing transmitters and receivers whose oscillators and clocks are not locked together.  For now we do not address the problem of fading, but we believe the correction of fading through equalization has the potential to also be addressed as an attention model, either jointly or subsequently to the transformations addressed here.

\subsection{Timing and Symbol-Rate Recovery}

Timing and symbol-rate recovery are relatively straightforward processes involving the re-sampling of the input signal at the correct starting offset and sampling increment.  This is very much akin to the extraction of visual pixels at the correct offset of a 1D Affine transformation, and so we treat it as such by directly leveraging the Affine transformation used in the image domain.  We represent our data as a 2D image, with a two rows containing I/Q and an N columns containing samples in time.

A full 2D Affine transformation allows for translation, rotation, and scaling in 2-dimensions given by a 2x6 element parameter vector.  To restrict this to 1-Dimensional translation and scaling in the time dimension, we can simply introduce the following mask \ref{eq:affinemask} and then readily use 2D Affine transform implementations from the image domain.  

\begin{equation} \label{eq:affinemask}
    \begin{bmatrix}
        \theta_0 & 0 & \theta_2 \\
        0 & \theta_1 & 0 
    \end{bmatrix}
\end{equation}

\subsection{Phase and Frequency Offset Recovery}

Phase and frequency offset recovery is a task which doesn't have an immediate analogue in the vision domain.  However this transform in signal processing is relatively straightforward.  We simply mix our signal with a complex sinusoid with the proper initial phase and frequency as defined by two new unknown parameters.

\begin{equation} \label{eq:rotation}
    y_n = x_n * exp(n \theta_3 + \theta_4)
\end{equation}

We directly implement this transform as a new layer in Keras (on top of Theano and Tensorflow), and cascade it before the Affine transform in the Transformer module of our network.

\subsection{Parameter Estimation}

The task of synchronization now becomes the task of parameter estimation of $\theta_i$ values passed into our transformer module.  We experimentally try a number of different neural network architectures for performing this parameter regression task, and ultimately introduce two new domain appropriate layers into Keras to help assist in their estimation.

\begin{figure*}[ht!]
  \centering
      \includegraphics[width=1.0\textwidth]{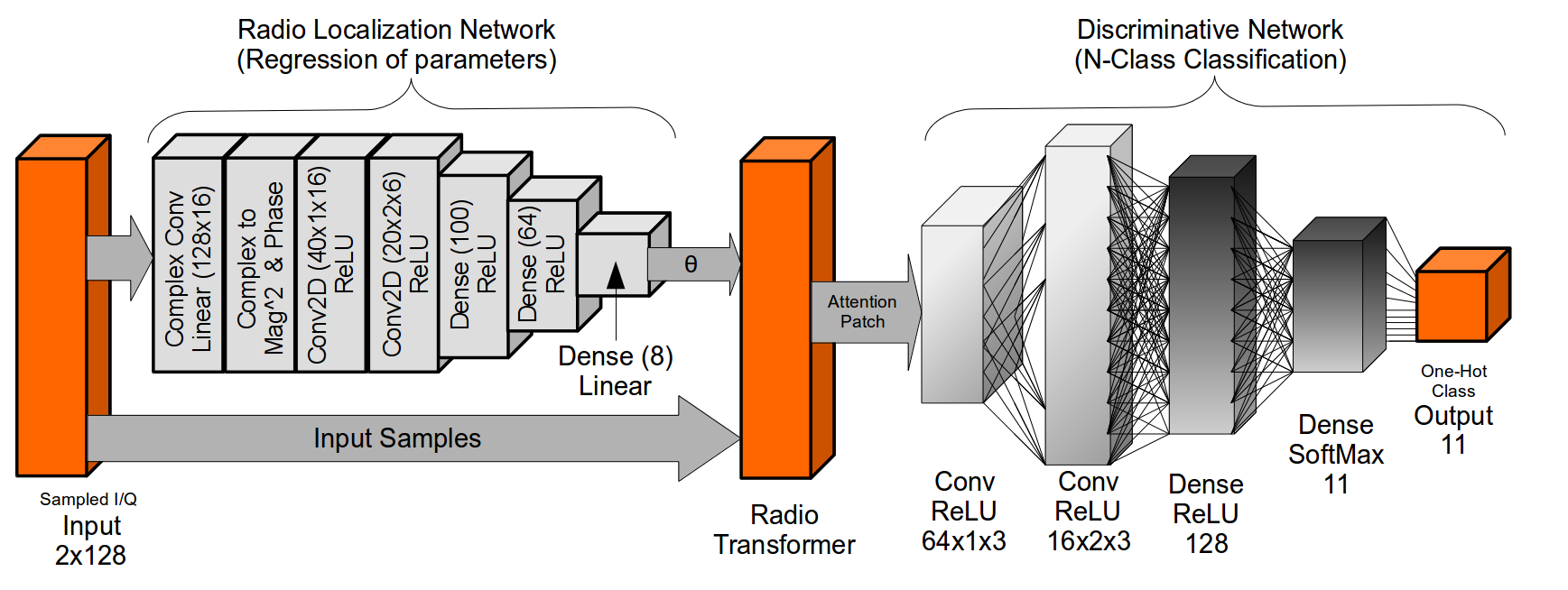}
        \caption{Radio Transformer Network Architecture}
    \label{fig:arch2}
\end{figure*}

\subsubsection{Complex Convolution 1D Layer}

Complex neural networks are not widely used and still faced some theoretical issues especially in automated differentiation, so we represent our signal as a two row 2D matrix with the real component in the first row and imaginary component in the second row.  In theory real valued convolutions in a neural network can learn the relationship between these components to some extent, but by introducing a complex convolution operation, we simplify the learning task and ensure that we learn a filter with the properties we are used to working with.

For a complex valued input vector X of size 2xN, we define a weight vector W of M complex filters each 2xK in length.  We may then compute the output for each of the k output values as.

\begin{equation} \label{eq:cpxconv}
    \begin{bmatrix}
        conv(X_{0,:}, W_{k,0,:}) - conv(X_{1,:}, W_{k,1,:}) \\
        conv(X_{0,:}, W_{k,1,:}) + conv(X_{1,:}, W_{k,0,:})
    \end{bmatrix}
\end{equation}

This allows us to leverage existing, highly optimized real convolution operations and obtain a differentiable operation which can be trained with back-propagation.

\subsubsection{Complex to Power and Phase}

Creating a differentiable Cartesian to Polar operation which makes it easier for the network to operate directly on input phase and magnitude is slightly more involved.  We compute magnitude squared simply as $m_n = pow(x_{0,n},2)+pow(x_{1,n},2)$, but for phase computation we use a simplified and differentiable approximation of atan2 without conditionals implemented in Keras on top of Theano and Tensorflow.  

\begin{verbatim}
z = Xq/K.clip(K.abs(Xi),1e-3,1e6)
z = z*K.sign(xi)
zd1 = 1+0.28*K.pow(z,2)
z1 = z/zd1 + PI*K.sign(Xq)
zd2 = K.pow(z,2) + 0.28
z2 = z/zd2 + (K.sign(Xq)-1)*0.5*PI
zc = K.abs(z) - 1
atan2 = (K.sign(zc)-1)*(-0.5)*z1 + 
        (K.sign(zc)+1)*0.5*z2
\end{verbatim}

\subsubsection{Network Architecture}

We evaluate dozens of localization network architectures with slight variation between dense connections, convolutional layers, and complex-convolutional layers, using various activation functions, and achieve our best performance using the composite network shown in figure \ref{fig:arch2}.  This uses both the complex convolutional layer and the complex to polar layers within the localization network and an identical discriminative network to the one we used without attention in front of it for comparison.

\subsubsection{Training Details}

We train this network using Keras \cite{keras} on top of Theano \cite{theano} and TensorFlow \cite{tensorflow} using an NVidia Geforce Titan X inside a Digits Devbox.  We use dropout \cite{dropout} of 0.5 between each layer for regularization, and the Adam \cite{adam} method of stochastic gradient descent to fit network parameters on our training set.

We train with a batch size of 1024 and an initial learning rate of 0.001.  We train for roughly 350 epochs, reducing our learning rate by half each validation loss stops decreasing.  Training takes about 3 hours on a Titan X GPU, but feed-forward evaluation or signal classification takes less than 10ms.  

\section{Data-set and Methodology}

For our classifier performance evaluation, in this work and in prior work, we leverage the RadioML.com 2016.04C open source dataset and perform at 60/40 split between train and test sets.  This consists of 11 modulations (8 digital and 3 analog) at varying SNR levels, with random walk simulations on center frequency, sample clock rate, sample clock offset, and initial phase, as well as limited multi-path fading.  We believe it is critically important to test with real channel effects early to ensure realistic assumptions early in our models.  This dataset is labeled with SNR and Modulation-Type, when performing supervised training we use only the modulation-type labels of the training set, and evaluate the classification accuracy performance at each SNR label step for the test set.

\begin{figure}[ht!]
  \centering
      \includegraphics[width=0.5\textwidth]{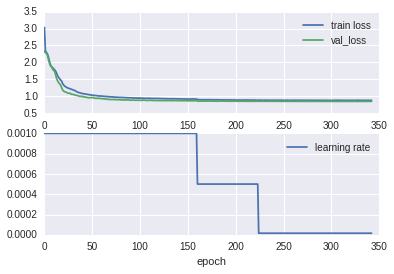}
        \caption{Training Loss and Learning Rate}
    \label{fig:loss}
\end{figure}

Training and validation loss along with learning rate are shown throughout the trainin in figure \ref{fig:loss}.

\section{Classification Performance}

\begin{figure}[ht!]
  \centering
      \includegraphics[width=0.5\textwidth]{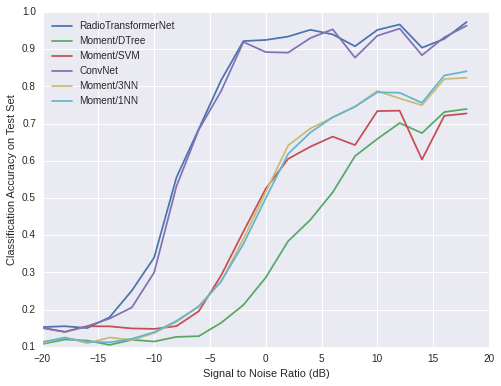}
        \caption{Radio Transformer Network Performance}
    \label{fig:perf2}
\end{figure}

Evaluating the performance of the RTN on the test set, we obtain slightly increased performance over the model without attention.  Similar accuracies are obtained at slightly lower SNR values (~1dB) and high SNR performance is slightly improved and more stable as shown in figure \ref{fig:perf2}.

We suspect the complexity of the discriminative network could now be reduced due to lower complexity of the normalized signal but we do not investigate this work further here for fair comparison of the same discriminative network.

Performance of the convolutional neural network without attention is also improved from our prior work by increasing dropout and better learning rate policy to match that used in our RTN training.  This is reflected in figure \ref{fig:perf2}.

\begin{figure}[ht!]
  \centering
      \includegraphics[width=0.5\textwidth]{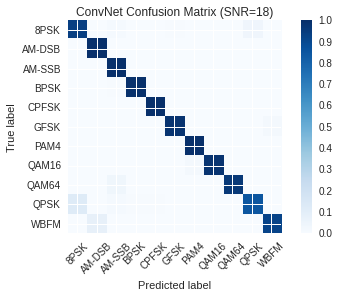}
        \caption{Performance of the RTN at High SNR (18dB)}
    \label{fig:highconf}
\end{figure}

\section{Attention Learning Performance}

\begin{figure}[ht!]
  \centering
      \includegraphics[width=0.5\textwidth]{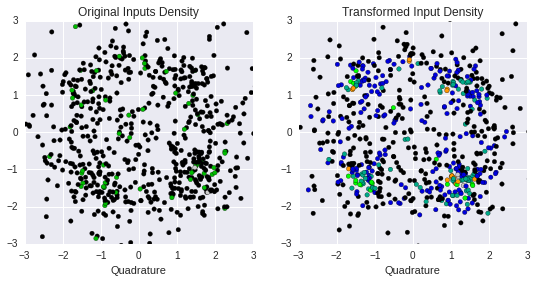}
        \caption{Density Plots of Pre- and Post-Transformed Input Constellations}
    \label{fig:density}
\end{figure}

We have shown that classification performance is improved while using a radio localization network to extract normalized patches on the dataset.  However, we can not only look at classification performance, it is also interesting to look at the radio sample data before and after transformation to observe any normalization that has occurred.  It is difficult to visualize exactly what has occured here, looking at time domain data does not yield clean, obvious performance improvement on the data.  Upon attempting to plot an eye-diagram of the QPSK-class signals, it is clear that synchronizaiton is at this point still horribly noisy and partial.   However, if we plot the the constellation density for 50 test examples over a range of 20 time samples each, shown in figure \ref{fig:density}, we can start to see a bit more density forming around the constellation points vs what we started with, which is a good sign to start with.

Clearly much work needs to be done to improve and quantify synchronization performance, in-fact we have no real reason to expect perfect synchronization from a classification task, just enough normalization to make things easier on the discriminative network.  We will continue to investigate this area, and additional tasks other than modulation recognition which may improve the synchronization properties for demodulation beyond what has been achieved here.

\section{Conclusions}

By developing a feed forward model for radio attention, we have demonstrated that we can effectively learn to synchronize using deep convolutional neural networks with domain specific transforms and layer configurations.  Normalizing out time, time-dilation, frequency and phase offsets using learned estimators does effectively improve our modulation classification performance and requires no expert knowledge about the signals of interest to train.  

While the training complexity of such a network is high, the feed-forward execution of it is actually quite compact and viable for real-world use and deployment.  Platforms such as the TX1 with highly parallel, low clock rate GPGPU architectures further enable the low-SWaP deployment of these algorithms for which they are exceptionally well suited.

\section*{Acknowledgments}

The authors would like to thank the Bradley Department of Electrical and Computer Engineering at the Virginia Polytechnic Institute and State University,the Machine Learning \& Perception Group, the Hume Center, and DARPA all for their generous support in this work.

This research was developed with funding from the Defense Advanced Research Projects Agency's (DARPA) MTO Office under grant HR0011-16-1-0002. The views, opinions, and/or findings expressed are those of the author and should not be interpreted as representing the official views or policies of the Department of Defense or the U.S. Government.

\nocite{clancy2007applications}

\printbibliography

\end{document}